\documentclass{article}

\usepackage{microtype}
\usepackage[final]{graphicx}
\usepackage{subcaption}
\usepackage{booktabs}
\usepackage{hyperref}
\usepackage{fvextra}
\usepackage[most]{tcolorbox}
\usepackage{multirow}
\usepackage{float}
\usepackage{placeins}

\usepackage[preprint]{icml2026}
\usepackage{enumitem}

\usepackage{amsmath}
\usepackage{amssymb}
\usepackage{mathtools}
\usepackage{amsthm}
\usepackage[capitalize,noabbrev]{cleveref}

\newcommand{\method}{\textsc{TrustMargin}}

\newcommand{\mprior}{M_{\mathrm{prior}}}
\newcommand{\mbind}{M_{\mathrm{bind}}}
\newcommand{\mcosa}{M_{\mathrm{TrustMargin}}}
\newcommand{\sparse}{\method}

\tcbset{
  cosaPrompt/.style={
    colback=gray!8,
    colframe=black!45,
    boxrule=0.4pt,
    arc=1pt,
    left=3pt,
    right=3pt,
    top=3pt,
    bottom=3pt,
    before skip=5pt,
    after skip=5pt
  }
}

\icmltitlerunning{\sparse}

\makeatletter
\renewcommand\section{\@startsection{section}{1}{\z@}{1.0ex}{0.55ex}{\normalfont\large\bfseries}}
\renewcommand\subsection{\@startsection{subsection}{2}{\z@}{0.8ex}{0.35ex}{\normalfont\normalsize\bfseries}}
\renewcommand\paragraph{\@startsection{paragraph}{4}{\z@}{0.45ex}{-0.55em}{\normalfont\normalsize\bfseries}}
\makeatother

\begin{document}

\twocolumn[
\begin{center}
{\LARGE\bfseries TrustMargin: Training-Free Arbitration between Parametric Memory and Retrieved Evidence in Large Language Models\par}
\vspace{0.22in}

\setlength{\tabcolsep}{0pt}
\begin{tabular}{@{}c@{\hspace{0.03\textwidth}}c@{\hspace{0.03\textwidth}}c@{}}
\parbox{0.30\textwidth}{\centering
{\large Jingyan Xu\textsuperscript{*}}\\
\texttt{\small 2309117485@emails.bjut.edu.cn}\\
DCST, Peking University\\
Beijing, China}
&
\parbox{0.30\textwidth}{\centering
{\large Hong Shi\textsuperscript{*}}\\
DCST, Peking University\\
Beijing, China}
&
\parbox{0.30\textwidth}{\centering
{\large Yi Shan\textsuperscript{*}}\\
DCST, Peking University\\
Beijing, China}
\\
\noalign{\vskip 0.16in}
\parbox{0.30\textwidth}{\centering
{\large Penghui Liu\textsuperscript{*}}\\
DCST, Peking University\\
Beijing, China}
&
\parbox{0.30\textwidth}{\centering
{\large Yunhao Bai}\\
DCST, Peking University\\
Beijing, China}
&
\parbox{0.30\textwidth}{\centering
{\large Ningyuan Li\textsuperscript{*}}\\
DCST, Peking University\\
Beijing, China}
\\
\noalign{\vskip 0.16in}
&
\parbox{0.30\textwidth}{\centering
{\large Xueyang Liu\textsuperscript{\ensuremath{\dagger}}}\\
\texttt{\small liuxueyang@pku.edu.cn}\\
DCST, Peking University\\
Beijing, China}
&
\end{tabular}
\end{center}

\vskip 0.3in
]

\begingroup
\makeatletter
\renewcommand{\@makefntext}[1]{\noindent #1}
\makeatother
\renewcommand{\thefootnote}{}
\footnotetext{%
\begin{tabular}{@{}p{\columnwidth}@{}}
\textsuperscript{*}Work done during their internships at Peking University.\\
\textsuperscript{\ensuremath{\dagger}}Corresponding author\\
\textsuperscript{1}We have open-sourced all the code and data at: \url{https://github.com/mojixu/TrustMargin.git}.
\end{tabular}}
\endgroup
\makeatletter
\global\icml@noticeprintedtrue
\makeatother

\begin{abstract}
Large language models answer knowledge-intensive questions using both parametric memory and retrieved evidence, but neither source is uniformly reliable. Retrieval can fill knowledge gaps, yet distracting passages may override correct closed-book answers. We study this post-generation conflict as answer-level source arbitration: given Direct and RAG answers from the same frozen model, decide which source to trust. We propose \method{}, a training-free, plug-and-play arbitration layer that scores the two existing candidates with the model's own likelihoods. It combines a parametric-prior margin, which tests whether memory accepts the retrieved answer, with an evidence-binding margin, which discounts passage-only salience and measures question-specific support. \method{} selects between Direct and RAG without fine-tuning, external judges, or additional generation. Across \textsc{2WikiMQA} and \textsc{CWQA} with three LLaMA scales, \method{} consistently improves over Direct generation and BM25-RAG, recovers part of the Direct/RAG oracle gap, and generalizes to multiple training-free RAG pipelines.
\end{abstract}

\section{Introduction}
\label{sec:intro}

Retrieval-augmented generation (RAG) has become a standard way to improve large language models (LLMs) on knowledge-intensive tasks \citep{lewis2020rag,guu2020realm,borgeaud2022retro,izacard2021fid,izacard2023atlas}.
By conditioning generation on external passages, RAG can supply long-tail or updated facts without changing model parameters.
Yet retrieval also introduces a reliability problem that is often hidden behind aggregate accuracy.
For some questions, retrieved passages supply the missing evidence; for others, they introduce distractors or long-context burden that pull the model away from a correct closed-book answer \citep{shi2023distracted,liu2024lost,levy2024same,yoran2023robust}.
A system that has access to both parametric memory and retrieved evidence therefore still faces a source decision: which one should dominate the final answer?

\begin{figure}[!tbp]
\centering
\includegraphics[width=\columnwidth,trim=8pt 6pt 8pt 6pt,clip]{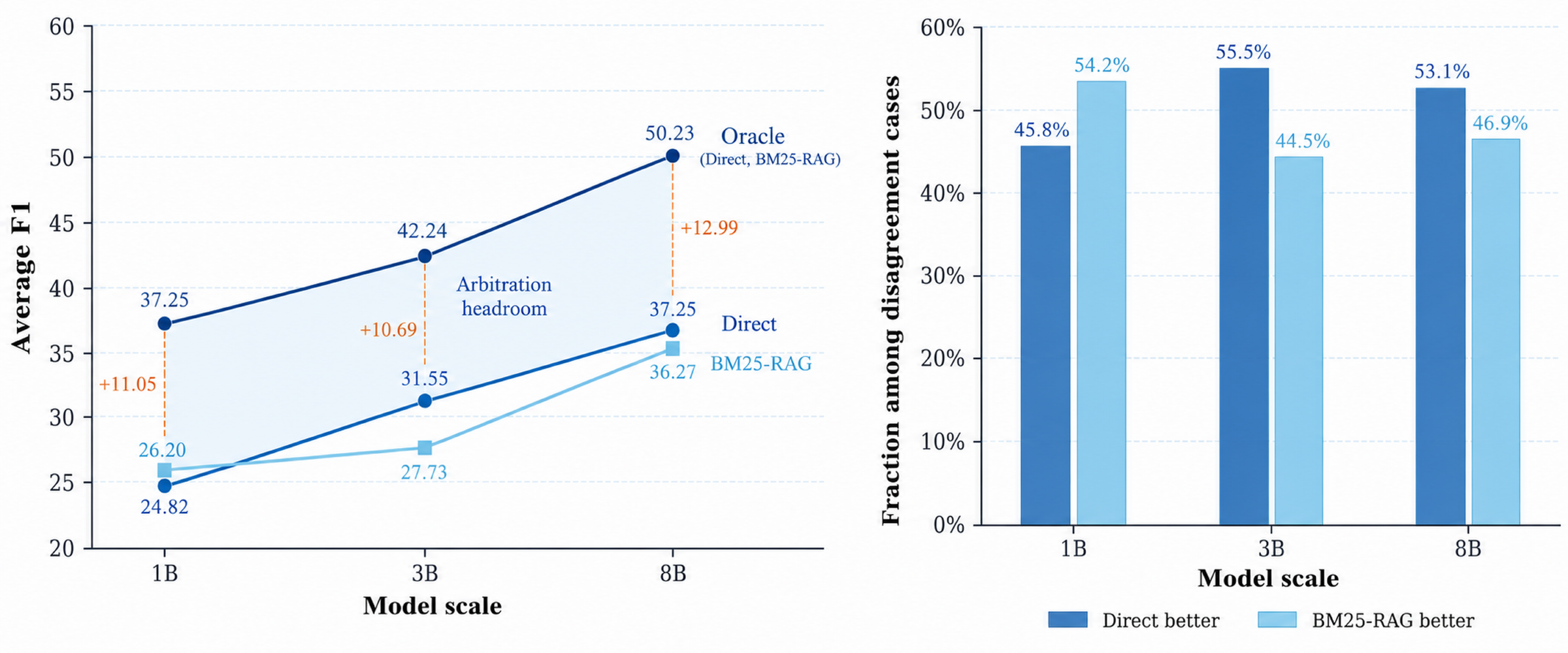}
\vspace{-6pt}
\caption{Motivation for answer-level source arbitration.
The Direct/RAG oracle exposes substantial candidate-set headroom across model scales, while disagreement cases are split between Direct-better and BM25-RAG-better examples.
The bottleneck is therefore not whether retrieval is globally useful, but when the retrieved answer should override parametric memory.}
\label{fig:motivation}
\vspace{-6pt}
\end{figure}

Figure~\ref{fig:motivation} shows that this conflict is not a rare failure mode.
The oracle that selects the better answer between Direct generation and BM25-RAG remains far above either individual source, meaning that many examples already contain a useful answer in the two-candidate set.
At the same time, the disagreement split is bidirectional: Direct is better for many examples, and BM25-RAG is better for many others.
A fixed source policy is therefore inadequate.
Always trusting retrieval is brittle, but always preserving memory leaves retrieved evidence unused.
The missing component is therefore not another global preference for retrieval or memory, but an instance-level rule for deciding when retrieved evidence should override parametric memory.

Most existing methods route retrieval rather than arbitrate between completed answers.
Adaptive retrieval methods decide whether or when to retrieve \citep{mallen2023trust,jiang2023flare,asai2024selfrag,jeong2024adaptive,su2024dragin}; retrieval-aware training, ranking, compression, or black-box augmentation methods modify how evidence is selected or consumed \citep{lin2023radit,zhang2024raft,yu2024rankrag,xu2024recomp,shi2024replug}; and parameter-level retrieval or editing methods inject external knowledge into weights or plug-in modules \citep{chen2023reckoning,hardt2024ttt,su2025prag,meng2022rome,meng2023memit}.
These approaches are powerful, but they change the generation pipeline through training, control tokens, online adaptation, or model-side parameterization.
They do not directly address the common setting in which both a closed-book answer and a RAG answer have already been produced.

We study this complementary setting as \emph{answer-level source arbitration}.
Given a question $q$ and retrieved passage set $P$, suppose the same frozen model produces a Direct answer $y_D=G_\theta(q)$ and a RAG answer $y_R=G_\theta(q,P)$.
The goal is to choose between $y_D$ and $y_R$ without modifying the model, changing the retriever, training a judge, or generating a third answer.
Because the decision is posterior to generation, it can be replayed on cached predictions and attached to existing Direct/RAG systems as a plug-and-play reliability layer when both candidate sources are available.

A reliable retrieval override should satisfy two complementary checks.
The first is \emph{prior compatibility}: under the question-only prompt, the frozen model should not strongly prefer its Direct answer over the RAG answer.
The second is \emph{question-context binding}: the answer should gain support from the interaction between the question and retrieved passages, rather than from passage-only salience.
Both signals can be measured with normalized teacher-forced likelihoods over the two already available answers.

We instantiate these checks in \method{}, a sparse training-free source arbitration rule.
For each question, \method{} computes a parametric-prior margin $\mprior$ under the question-only prompt and an evidence-binding margin $\mbind$ by contrasting question-plus-context likelihoods with context-only likelihoods.
It then forms a trust score
\begin{equation}
M = \mprior + \lambda_{\mathrm{bind}}\mbind .
\label{eq:cosa_score}
\end{equation}
The RAG answer is selected only when $M>\tau$; otherwise, the Direct answer is preserved.
In the main experiments, we use $\lambda_{\mathrm{bind}}=0.5$ and $\tau=-1.5$.
The method is intentionally restrictive: it does not create new candidates, but decides whether the RAG answer has enough support to replace the Direct answer.

We evaluate \method{} on \textsc{2WikiMQA} and \textsc{CWQA} with LLaMA-3.2-1B, LLaMA-3.2-3B, and LLaMA-3.1-8B.
Across all six model-dataset settings, \method{} improves over both Direct generation and BM25-RAG while keeping candidate generation fixed.
Ablations show that prior compatibility and evidence binding are complementary: the former guards against retrieval-induced drift, while the latter identifies cases where retrieval provides question-specific support.
Diagnostics further show that \method{} recovers part of the Direct/RAG oracle gap, becomes more conservative under retrieval corruption, and remains useful for several training-free RAG pipelines when their RAG-side candidates provide useful diversity.

In summary, this paper makes three contributions:
\begin{itemize}[leftmargin=*, nosep]
\item We formulate \emph{answer-level source arbitration} as a distinct reliability problem for RAG systems: after both closed-book and retrieval-augmented answers have been produced, decide which source should be trusted, rather than whether retrieval should be invoked.
\item We propose \method{}, a training-free and judge-free arbitration rule based on pairwise likelihood margins. The parametric-prior margin acts as a memory-side guardrail, while the evidence-binding margin discounts passage-only salience to measure question-specific support.
\item We show that \method{} consistently improves F1 and EM over Direct generation and BM25-RAG across datasets and model scales, and further demonstrate through diagnostics that it recovers Direct/RAG complementarity, resists retrieval corruption, and can be plugged into several training-free RAG pipelines when their candidates provide useful diversity.
\end{itemize}

\section{Related Work}
\label{sec:related}

\paragraph{Parametric and non-parametric knowledge.}
LLMs store factual knowledge in their parameters and can answer many questions without external context \citep{petroni2019language,roberts2020how,brown2020language}.
RAG and nearest-neighbor methods complement this memory with evidence retrieved at inference time \citep{khandelwal2020knnlm,lewis2020rag,guu2020realm,ram2023incontext}.
Prior work shows that these sources are complementary rather than globally ordered \citep{mallen2023trust}.
\method{} moves this complementarity to the answer level: once Direct and RAG answers both exist, it decides which source to trust for the current instance.

\paragraph{RAG architectures and evidence use.}
RAG systems improve knowledge-intensive generation through sparse or dense retrieval, retrieval pretraining, reader architectures, and context-based evidence use \citep{karpukhin2020dpr,borgeaud2022retro,izacard2021fid,izacard2023atlas}.
Other work improves query formulation, black-box augmentation, or context compression \citep{mao2021gar,shi2024replug,xu2024recomp}.
These methods aim to produce better retrieved-context answers.
\method{} is orthogonal: it leaves the RAG pipeline unchanged and decides whether its answer should override the closed-book answer.

\paragraph{Adaptive and reasoning-time retrieval.}
Adaptive RAG methods decide whether, when, or how to retrieve during generation \citep{jiang2023flare,asai2024selfrag,jeong2024adaptive,su2024dragin,chen2026dtr}.
Retrieval can also be interleaved with reasoning traces for multi-step QA and long-horizon generation \citep{wei2022cot,yao2023react,trivedi2023ircot,wang2024rat}.
These methods intervene before or during generation.
\method{} operates after generation and can be attached to their outputs as an arbitration layer.

\paragraph{Parameter-level knowledge injection and editing.}
A parallel line of work studies factual knowledge in model weights and how to edit or adapt it \citep{geva2021transformer,dai2022knowledge,meng2022rome,meng2023memit}.
Model editing and test-time adaptation modify weights or attach parameter-efficient modules \citep{mitchell2022mend,mitchell2022serac,chen2023reckoning,hardt2024ttt,hu2022lora}, while Parametric RAG retrieves document-level parameter modules \citep{su2025prag}.
\method{} does not inject knowledge or update parameters; it uses the frozen model's likelihoods to arbitrate between two existing answers.

\paragraph{Distracting and weakly bound context.}
Retrieved context can help factuality but also degrade reasoning when passages are irrelevant, overly long, or only topically related \citep{shi2023distracted,liu2024lost,levy2024same,yoran2023robust}.
Training-free decoding methods reduce distractibility by emphasizing more reliable evidence distributions \citep{qiu2025entropy}.
\method{} addresses the same problem at the source-selection level: its binding margin subtracts context-only likelihood from question-plus-context likelihood, turning raw passage salience into a question-specific evidence signal.

\section{Methodology}
\label{sec:method}

\begin{figure*}[!t]
\centering
\includegraphics[width=0.98\textwidth]{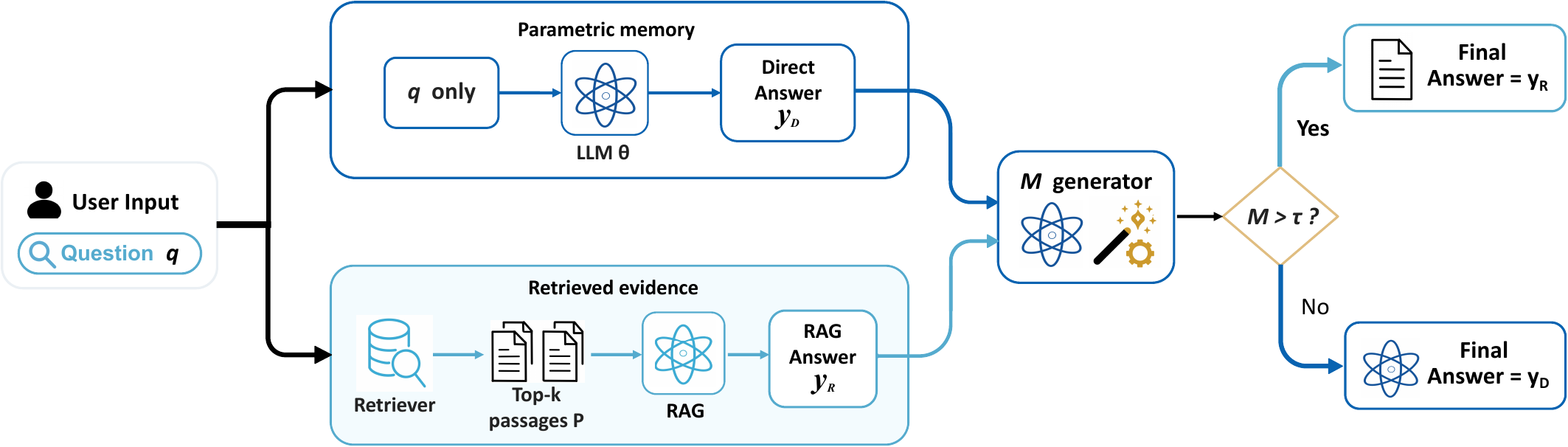}
\caption{Overview of the \method{} framework.
The same frozen LLM produces a Direct answer $y_D$ from the question alone and a RAG answer $y_R$ from the question plus retrieved passages.
The $M$-generator scores both candidates and returns a trust score $M$.
It does not generate a new answer; it only evaluates the existing Direct and RAG candidates.
The final decision is sparse: select the RAG answer only when $M>\tau$; otherwise preserve the Direct answer.}
\label{fig:tm_overview}
\end{figure*}

\begin{figure}[!htb]
\centering
\includegraphics[width=\columnwidth]{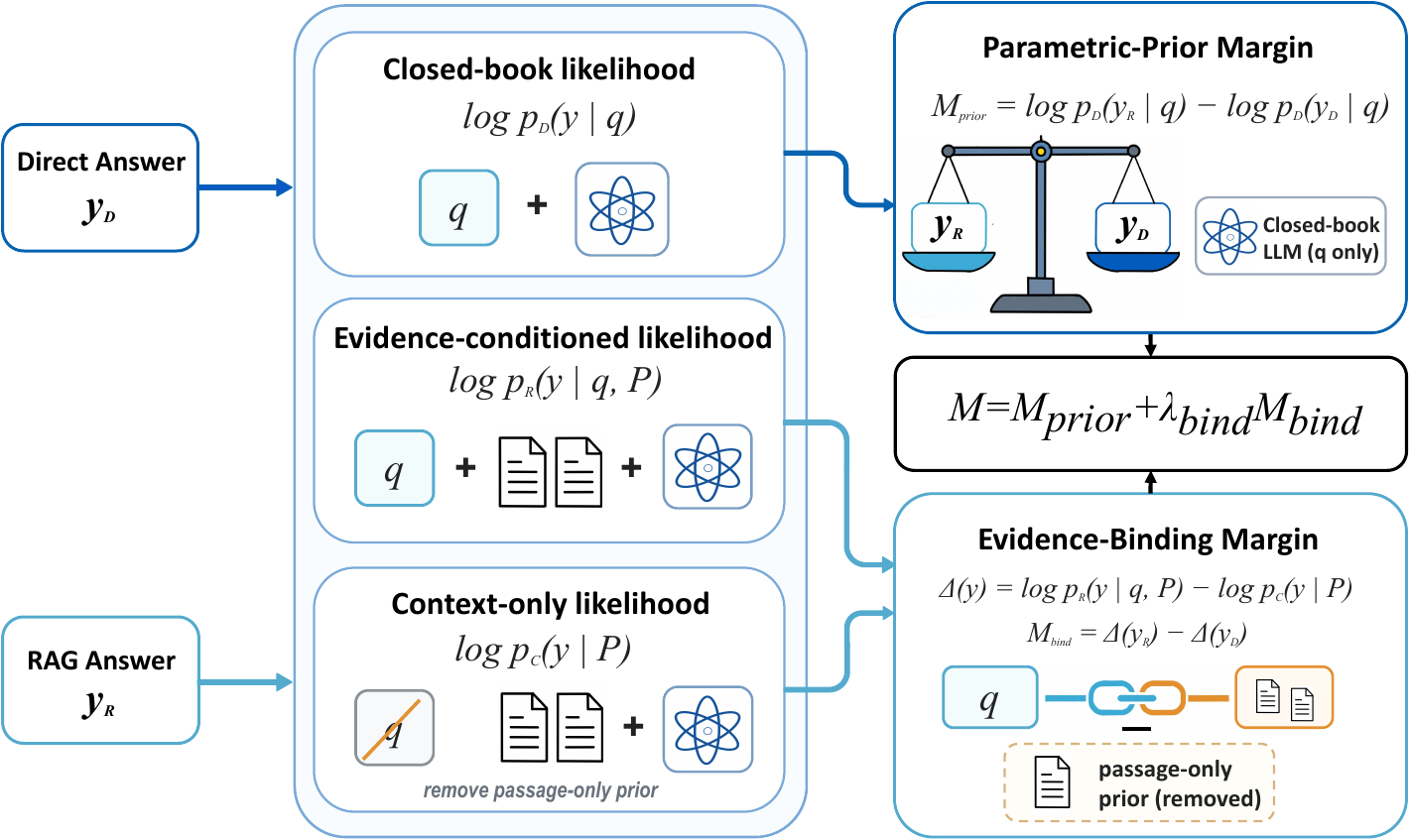}
\caption{Detailed view of the $M$-generator.
Both candidate answers are scored under closed-book, evidence-conditioned, and context-only likelihood views.
The parametric-prior margin compares the Direct and RAG answers under the question-only prompt.
The evidence-binding margin subtracts passage-only salience from evidence-conditioned support, then compares the two candidates.
The final trust score is $M=\mprior+\lambda_{\mathrm{bind}}\mbind$.}
\label{fig:tm_generator}
\end{figure}

This section defines \method{}, a training-free rule for selecting between an existing Direct answer and an existing RAG answer.
Figure~\ref{fig:tm_overview} shows the pipeline: the frozen LLM produces $y_D$ from the question alone and $y_R$ from retrieved passages; the $M$-generator scores both candidates and applies a threshold decision.
Figure~\ref{fig:tm_generator} expands the $M$-generator into three likelihood views and two margins.

\subsection{Answer-Level Source Arbitration}
\label{sec:problem_formulation}

Let $q$ be a question and $P=\{p_i\}_{i=1}^{20}$ be the top-20 passages retrieved by BM25 \citep{robertson2009bm25}.
A frozen LLM $G_\theta$ produces two answers under aligned prompts:
\begin{equation}
\label{eq:candidates}
y_D = G_\theta(q), \qquad y_R = G_\theta(q,P).
\end{equation}
The task is to select $\hat{y}\in\{y_D,y_R\}$ without changing the model, retriever, generation prompts, or candidate set.
The decision is posterior to generation: it does not trigger retrieval, train a verifier, or generate another answer.

\subsection{The $M$-Generator: Three Likelihood Views}
\label{sec:m_generator}

The $M$-generator scores both candidates under three teacher-forced, length-normalized likelihood views:
\begin{align}
\ell_D(y) &= \frac{1}{|y|}\log p_\theta(y\mid q), \\
\ell_R(y) &= \frac{1}{|y|}\log p_\theta(y\mid q,P), \\
\ell_C(y) &= \frac{1}{|y|}\log p_\theta(y\mid P).
\end{align}
Here $\ell_D$ probes the model's closed-book preference, $\ell_R$ measures evidence-conditioned support, and $\ell_C$ estimates passage-only salience after the question is removed.

\subsection{Parametric-Prior Margin}
\label{sec:prior}

The prior margin checks whether the frozen model accepts the RAG answer before seeing retrieved passages:
\begin{equation}
\label{eq:prior}
\mprior = \ell_D(y_R)-\ell_D(y_D).
\end{equation}
A positive value means the RAG answer is already plausible under the question-only prompt; a strongly negative value means parametric memory prefers the Direct answer.
Thus, $\mprior$ acts as a memory-side guardrail against unsupported retrieval overrides.

\subsection{Evidence-Binding Margin}
\label{sec:binding}

Evidence-conditioned likelihood can confuse salience with support: a passage may mention an entity without making it the answer.
We remove this passage-only prior by defining
\begin{equation}
\label{eq:binding_score}
\Delta(y)=\ell_R(y)-\ell_C(y).
\end{equation}
and compare the two candidates:
\begin{equation}
\label{eq:binding_margin}
\mbind = \Delta(y_R)-\Delta(y_D).
\end{equation}
A large positive $\mbind$ means the RAG answer gains more support from the question-evidence interaction than the Direct answer does.

\subsection{Sparse Source Decision}
\label{sec:decision}

\method{} combines the two margins into
\begin{equation}
\label{eq:cosa_score_method}
\mcosa = \mprior + \lambda_{\mathrm{bind}}\mbind.
\end{equation}
The final answer is
\begin{equation}
\label{eq:decision}
\hat{y}=
\begin{cases}
y_R, & \mcosa>\tau, \\
y_D, & \mcosa\leq\tau.
\end{cases}
\end{equation}
We use $\lambda_{\mathrm{bind}}=0.5$ and $\tau=-1.5$ across all main experiments.
The nonzero threshold makes the rule conservative: retrieval must provide enough combined support before replacing the Direct answer.

\subsection{Interpretation}
\label{sec:interpretation}

The two margins encode different failure modes.
$\mprior$ asks whether memory accepts the retrieved answer; $\mbind$ asks whether evidence is question-bound rather than merely salient.
Their combination forms a compact boundary between preserving memory and trusting retrieval.
Because \method{} only selects between $y_D$ and $y_R$, it remains candidate-set bounded: it cannot fix examples where both answers are wrong.

\begin{table*}[!t]
\centering
\caption{
Main results on \textsc{2WikiMQA} and \textsc{CWQA}.
We compare \method{} with training-free baselines across three LLaMA model scales.
F1 and EM denote token-level F1 and exact match, respectively.
``Average'' is the arithmetic mean over the two datasets.
All retrieval-based methods use the top-20 BM25 passages unless otherwise specified.
The best result within each model block is bolded and the second-best result is underlined.
}
\label{tab:main_results}
\scriptsize
\setlength{\tabcolsep}{4pt}
\renewcommand{\arraystretch}{0.95}
\resizebox{0.86\textwidth}{!}{
\begin{tabular}{llcccccc}
\toprule
 & & \multicolumn{2}{c}{\textbf{\textsc{2WikiMQA}}} & \multicolumn{2}{c}{\textbf{\textsc{CWQA}}} & \multicolumn{2}{c}{\textbf{\textsc{Average}}} \\
\cmidrule(lr){3-4} \cmidrule(lr){5-6} \cmidrule(lr){7-8}
\textbf{Model} & \textbf{Method} & \textbf{F1} & \textbf{EM} & \textbf{F1} & \textbf{EM} & \textbf{F1} & \textbf{EM} \\
\midrule
\multirow{7}{*}{\textbf{LLaMA3.2-1B}}
& Direct      & 22.17 & 16.10 & 27.46 & 21.10 & 24.82 & 18.60 \\
& BM25-RAG    & \underline{26.40} & 19.40 & 26.01 & 19.70 & 26.20 & 19.55 \\
& IRCoT       & 26.32 & 20.10 & 27.91 & 21.00 & 27.12 & 20.55 \\
& FLARE       & 22.77 & 16.60 & 28.74 & 21.50 & 25.75 & 19.05 \\
& CLeHe-RAG   & 26.14 & \textbf{20.40} & 30.49 & \underline{23.20} & 28.31 & \underline{21.80} \\
& DTR-RAG     & 26.26 & \textbf{20.40} & \underline{30.53} & \textbf{23.70} & \underline{28.40} & \textbf{22.05} \\
& \textbf{TrustMargin (Ours)} & \textbf{27.29} & \textbf{20.40} & \textbf{30.60} & \underline{23.20} & \textbf{28.95} & \underline{21.80} \\
\midrule
\multirow{7}{*}{\textbf{LLaMA3.2-3B}}
& Direct      & 24.27 & 17.90 & \underline{38.83} & \underline{30.60} & 31.55 & 24.25 \\
& BM25-RAG    & 25.33 & 18.80 & 30.14 & 23.00 & 27.73 & 20.90 \\
& IRCoT       & 27.45 & 20.70 & 30.03 & 23.90 & 28.74 & 22.30 \\
& FLARE       & 24.04 & 17.90 & 38.27 & 30.20 & 31.16 & 24.05 \\
& CLeHe-RAG   & 23.13 & 16.30 & 36.33 & 28.00 & 29.73 & 22.15 \\
& DTR-RAG     & \underline{27.65} & \underline{21.50} & 36.31 & 29.30 & \underline{31.98} & \underline{25.40} \\
& \textbf{TrustMargin (Ours)} & \textbf{30.29} & \textbf{22.90} & \textbf{40.73} & \textbf{31.60} & \textbf{35.51} & \textbf{27.25} \\
\midrule
\multirow{7}{*}{\textbf{LLaMA3.1-8B}}
& Direct      & 30.98 & 24.90 & 43.51 & 33.70 & 37.25 & 29.30 \\
& BM25-RAG    & \underline{34.30} & \underline{27.30} & 38.25 & 30.20 & 36.28 & 28.75 \\
& IRCoT       & 33.12 & 26.20 & 38.17 & 29.70 & 35.65 & 27.95 \\
& FLARE       & 31.61 & 24.90 & \underline{43.74} & \underline{34.10} & 37.68 & 29.50 \\
& CLeHe-RAG   & 27.13 & 22.90 & 40.63 & 33.20 & 33.88 & 28.05 \\
& DTR-RAG     & 33.67 & \underline{27.30} & 41.72 & 33.70 & \underline{37.70} & \underline{30.50} \\
& \textbf{TrustMargin (Ours)} & \textbf{39.48} & \textbf{32.70} & \textbf{45.74} & \textbf{36.10} & \textbf{42.61} & \textbf{34.40} \\
\bottomrule
\end{tabular}
}
\end{table*}

\section{Experimental Setup}
\label{sec:setup}

\subsection{Benchmarks and Metrics}
\label{sec:benchmarks_metrics}

We evaluate \method{} on \textsc{2WikiMQA} \citep{ho2020twowiki} and \textsc{CWQA} \citep{talmor2018cwq}.
\textsc{2WikiMQA} stresses multi-hop reasoning over Wikipedia-style evidence, where retrieval can supply missing facts but also partial contexts.
\textsc{CWQA} contains complex web questions where closed-book generation remains competitive, making harmful retrieval overrides visible.
Together, the benchmarks test whether an arbiter can use retrieval when it helps while preserving memory when retrieval misleads.

We report token-level F1, Exact Match (EM), and their arithmetic average across the two datasets.

\subsection{Baselines}
\label{sec:baselines}

We compare against training-free baselines covering fixed source policies and retrieval-time interventions.

\begin{itemize}[leftmargin=*, nosep]
    \item \textbf{Direct Generation.} The LLM answers without retrieved passages; this measures closed-book parametric memory.
    \item \textbf{BM25-RAG.} The LLM answers using the top-20 BM25 passages \citep{robertson2009bm25}; this always trusts retrieval once passages are provided.
    \item \textbf{IRCoT.} IRCoT interleaves retrieval with intermediate reasoning \citep{trivedi2023ircot}.
    \item \textbf{FLARE.} FLARE triggers retrieval during generation based on uncertainty \citep{jiang2023flare}.
    \item \textbf{CLeHe-RAG.} We use the training-free entropy-based decoding method of \citet{qiu2025entropy}, referred to as CLeHe-RAG in our implementation. This baseline improves evidence utilization before answer generation and tests whether \method{} remains useful when the RAG-side candidate is stronger than vanilla BM25-RAG.
    \item \textbf{DTR-RAG.} DTR-RAG is a training-free retrieval framework that adaptively decides when to retrieve and constructs evidence through a dual-path retrieval mechanism \citep{chen2026dtr}. It tests whether answer-level arbitration complements decoding-time retrieval control.
    \item \textbf{\method{} (Ours).} \method{} reuses the Direct and BM25-RAG candidates, scores them with the frozen LLM, and selects one answer with the two-margin rule.
\end{itemize}

For mechanism analysis, we evaluate single-margin variants using only $\mprior$ or only $\mbind$.
We use $\operatorname{Oracle}(\mathrm{Direct},\mathrm{RAG})$ only as a post-hoc candidate-set upper bound.
All methods share the same model family, BM25 backend, short-answer prompt style, and greedy decoding unless noted otherwise.
This shared protocol keeps comparisons focused on source use rather than retrieval depth, prompting, or decoding.
Mechanism variants operate on the same Direct/RAG candidate pair as the full method, isolating the contribution of each margin.

\subsection{Implementation Details}
\label{sec:implementation_details}

\paragraph{Models and retrieval.}
We instantiate all methods on three frozen instruction-tuned LLaMA models: LLaMA-3.2-1B-Instruct, LLaMA-3.2-3B-Instruct, and LLaMA-3.1-8B-Instruct \citep{touvron2023llama,grattafiori2024llama3}.
We use BM25 \citep{robertson2009bm25} implemented with Elasticsearch as the retrieval backend.
For each question, the retriever returns the top-20 Wikipedia passages, which are shared by BM25-RAG generation and all \method{} likelihood-scoring views.
This shared evidence pool ensures that differences between BM25-RAG and \method{} come from source arbitration rather than from retrieval depth or retriever choice.

\paragraph{Candidate generation and scoring.}
For every example, we first generate two candidate answers with the same frozen model: a Direct answer from the question-only prompt and a RAG answer from the question-context prompt.
Both candidates are generated once with greedy decoding and then kept fixed.
\method{} does not generate an additional answer.
It computes length-normalized teacher-forced likelihoods for the two candidates under three prompt views: question-only, question-context, and context-only.
These scores define the parametric-prior margin $\mprior$ and the evidence-binding margin $\mbind$.

\paragraph{Runtime and hyperparameters.}
Unless otherwise specified, all methods use greedy decoding with temperature $0$, \texttt{do\_sample=false}, \texttt{num\_beams=1}, maximum context length $2048$, maximum generation length $20$, and random seed $42$.
For all main experiments, \method{} uses a fixed binding weight $\lambda_{\mathrm{bind}}=0.5$ and arbitration threshold $\tau=-1.5$ across datasets and model scales.
The final trust score is
\[
\mcosa = \mprior + \lambda_{\mathrm{bind}}\mbind,
\]
and the selected answer is $y_R$ if $\mcosa>\tau$ and $y_D$ otherwise.
All experiments were run on a single NVIDIA A100-SXM4-80GB GPU.
Prompt templates and additional implementation details are provided in Appendix~\ref{app:prompt_templates}.

\section{Experimental Results}
\label{sec:results}

\subsection{Main Results}
\label{sec:main_results}

Table~\ref{tab:main_results} compares \method{} with training-free baselines across model scales and benchmarks.
Direct and RAG fail in different regions: retrieval often helps on \textsc{2WikiMQA}, while on \textsc{CWQA} it can override a reliable closed-book answer.
A fixed source policy is therefore brittle.

\method{} avoids this commitment by selecting between the two completed answers instance by instance.
On the 8B model, for example, it raises average F1 from 37.25 for Direct and 36.28 for BM25-RAG to 42.61.
Scaling does not eliminate the source decision; stronger models gain memory, but they also exploit useful evidence more effectively when retrieval is reliable.

IRCoT, FLARE, CLeHe-RAG, and DTR-RAG intervene before or during generation.
\method{} instead acts after generation as a reusable arbitration layer.
Its competitiveness shows that RAG reliability is also a source-decision problem.

\subsection{Ablation Studies}
\label{sec:ablation}

\paragraph{Necessity of dual-margin arbitration.}
Table~\ref{tab:ablation_compact} tests whether the two margins capture distinct trust signals.
Removing either one weakens the method.

\begin{table}[!t]
\centering
\scriptsize
\caption{
Compact margin ablation under the top-20 passage setting.
The ``Full'' column reports the absolute average F1/EM of \method{} over \textsc{2WikiMQA} and \textsc{CWQA}.
The ablation columns report performance drops relative to the full method after removing either the parametric-prior margin $\mprior$ or the evidence-binding margin $\mbind$.
Larger drops indicate that the removed margin contributes more to the full arbitration rule.
}
\label{tab:ablation_compact}
\setlength{\tabcolsep}{3.5pt}
\renewcommand{\arraystretch}{1.05}
\begin{tabular}{@{}lccc@{}}
\toprule
Model
& Full
& $\Delta$ w/o $\mprior$
& $\Delta$ w/o $\mbind$ \\
& Avg. F1/EM
& $\Delta$F1/EM
& $\Delta$F1/EM \\
\midrule
LLaMA-1B
& \textbf{28.95/21.80}
& $-$1.77/$-$1.25
& $-$0.53/$-$0.20 \\
LLaMA-3B
& \textbf{35.51/27.25}
& $-$5.06/$-$3.80
& $-$1.31/$-$1.05 \\
LLaMA-8B
& \textbf{42.61/34.40}
& $-$3.63/$-$2.90
& $-$2.23/$-$2.15 \\
\bottomrule
\end{tabular}
\end{table}

The prior margin $\mprior$ is the memory-side guardrail: without it, fluent RAG answers can override a strong closed-book belief.
The binding margin $\mbind$ is the evidence-side check: without it, the router becomes too conservative when retrieval supplies missing facts.
By subtracting context-only likelihood, $\mbind$ separates question-specific support from passage-only salience.
Together, the margins accept retrieval only when the answer is plausible under memory and bound to the question evidence.

\begin{figure*}[!t]
\centering
\includegraphics[width=0.88\textwidth]{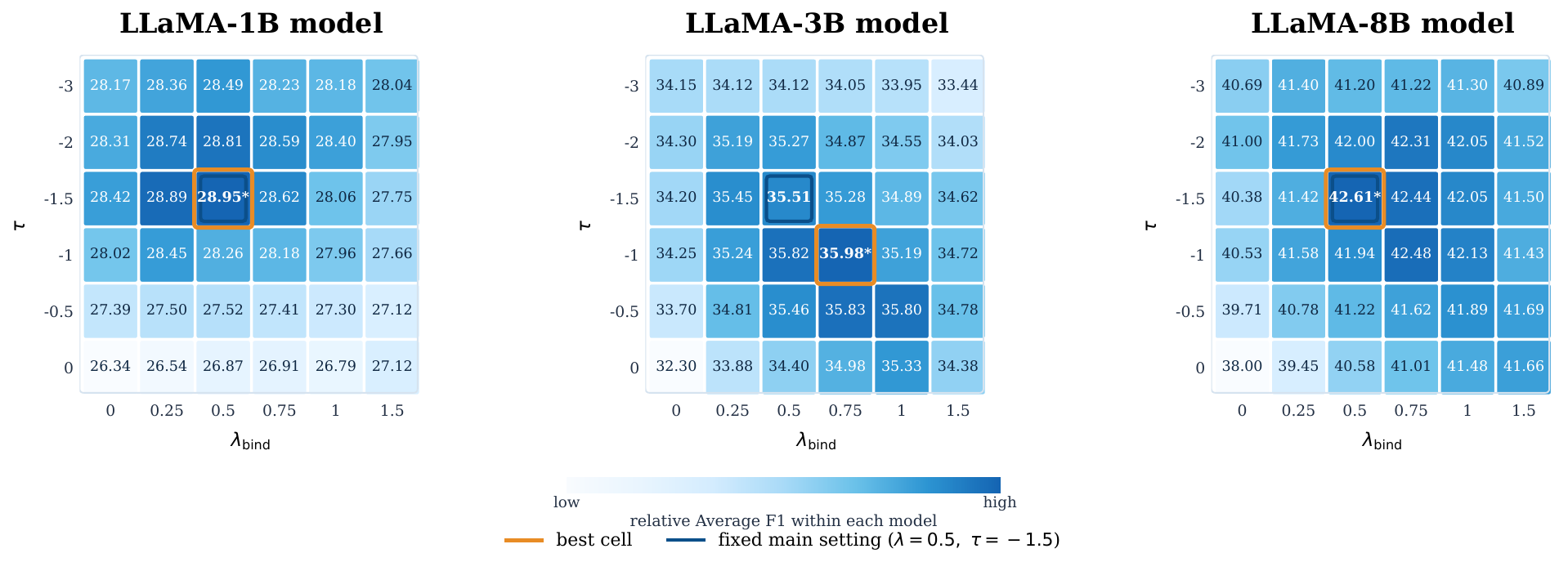}
\caption{
Hyperparameter robustness of \method{}.
Each cell reports average F1 over \textsc{2WikiMQA} and \textsc{CWQA} for a fixed pair of binding weight $\lambda_{\mathrm{bind}}$ and arbitration threshold $\tau$.
The purple box marks the fixed main setting $(\lambda_{\mathrm{bind}}=0.5,\tau=-1.5)$; the orange box marks the best cell for each model scale.
}
\label{fig:hparam_robustness}
\end{figure*}

\paragraph{Robustness to hyperparameters.}
Figure~\ref{fig:hparam_robustness} shows that the method is not driven by a fragile threshold.
Strong configurations occupy a broad region, justifying one fixed setting across datasets and model scales.

The heat map matches the intended sparse boundary.
A permissive threshold over-trusts retrieval; a conservative threshold collapses toward Direct generation.
The stable region lies between these extremes, where retrieval overrides memory only with sufficient combined support.

\subsection{Diagnostic Analysis}
\label{sec:diagnostic_analysis}

Since \method{} generates no new answer, its gains must come from complementarity already present in the candidate set.
We analyze oracle opportunity, disagreement behavior, recovered gain, retrieval-noise robustness, and compatibility with other training-free RAG pipelines.

\paragraph{Oracle gap and source arbitration.}
The Direct/RAG oracle selects the better candidate answer for each example.
For consistency with the main results and motivation analysis, $\operatorname{Oracle}(D,R)$ is computed over the same Direct and BM25-RAG candidates reported in Table~\ref{tab:main_results}.
It is not deployable, but it measures the value of perfect source selection.
As shown in Table~\ref{tab:diagnostic_arbitration_compact}, the candidate-set oracle remains substantially above the stronger single source, leaving an average F1 gap of roughly 10.7--13.0 points across model scales.

\begin{table}[!t]
\centering
\scriptsize
\setlength{\tabcolsep}{2.6pt}
\caption{
Source-arbitration diagnostics averaged over \textsc{2WikiMQA} and \textsc{CWQA} under the unified candidate-set definition.
$D$ and $R$ denote the Direct and BM25-RAG answers.
$\operatorname{Oracle}(D,R)$ selects the better candidate using the gold answer.
Oracle Gap is $\operatorname{Oracle}(D,R)$ minus the stronger single source, and Gap Closed is the percentage of this gap recovered by \method{}.
Source-selection rates are computed on strict disagreement cases.
}
\label{tab:diagnostic_arbitration_compact}
\resizebox{\columnwidth}{!}{%
\begin{tabular}{lcccc}
\toprule
Model & \shortstack{Oracle\\Gap} & \shortstack{Gap\\Closed} & $D{>}R{\to}D$ & $R{>}D{\to}R$ \\
 & F1 / EM & F1 / EM & \% & \% \\
\midrule
LLaMA-1B & 11.05 / 8.65 & 24.89 / 26.01 & 60.59 & 66.94 \\
LLaMA-3B & 10.69 / 8.90 & 37.04 / 33.71 & 65.77 & 77.74 \\
LLaMA-8B & 12.98 / 11.05 & 41.29 / 46.15 & 60.00 & 76.39 \\
\bottomrule
\end{tabular}
}
\end{table}

\method{} recovers 24.89--41.29\% of the F1 oracle gap and 26.01--46.15\% of the EM oracle gap.
A sizeable fraction of the answer-quality gap therefore lies in choosing between existing candidates rather than generating new ones.
The unrecovered gap marks the limit of the setting: when both candidates are wrong, a two-candidate selector cannot recover the gold answer.

\paragraph{Alignment with oracle-preferred sources.}
Disagreement cases are where arbitration matters.
Table~\ref{tab:diagnostic_arbitration_compact} shows the expected directional pattern: \method{} selects Direct more often when Direct is better, and RAG more often when RAG is better.
The margins therefore capture source reliability rather than a global preference for retrieval or memory.
Residual errors cluster near the boundary, where both candidates are plausible or retrieved entities are topical but weakly bound to the question.

\begin{figure}[!htb]
\centering
\includegraphics[width=0.86\columnwidth]{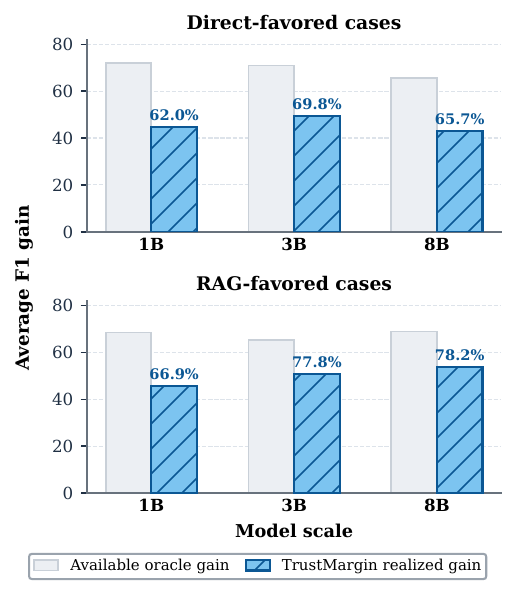}
\caption{
Disagreement recovery analysis.
Direct-favored and RAG-favored cases denote disagreement cases where Direct or BM25-RAG has higher F1, respectively.
Gray bars show the available oracle F1 gain from perfect Direct/RAG source selection, while blue bars show the gain realized by \method{}.
Percentages above blue bars report realized gain divided by available oracle gain.
}
\label{fig:disagreement_recovery}
\end{figure}

\paragraph{Disagreement Recovery Analysis.}
Selection accuracy ignores the size of each missed opportunity.
We therefore measure, separately for Direct-favored and RAG-favored cases, the fraction of available oracle F1 gain realized by \method{}.
This recovery metric differs from the source-selection rate in Table~\ref{tab:diagnostic_arbitration_compact}: selection rate asks whether the preferred source is chosen, while recovery measures realized F1 gain.
Figure~\ref{fig:disagreement_recovery} shows that \method{} converts Direct/RAG complementarity into answer quality in both regimes.

Recovery is higher on RAG-favored cases than on Direct-favored cases.
When retrieval supplies missing evidence, the binding margin often gives a strong positive signal for RAG.
Direct-favored cases are harder: misleading passages may contain topical entities that remain locally plausible.
Such weakly bound entities account for much of the remaining gap to the oracle.

\begin{figure}[!tb]
\centering
\includegraphics[width=\columnwidth,trim=10pt 8pt 10pt 8pt,clip]{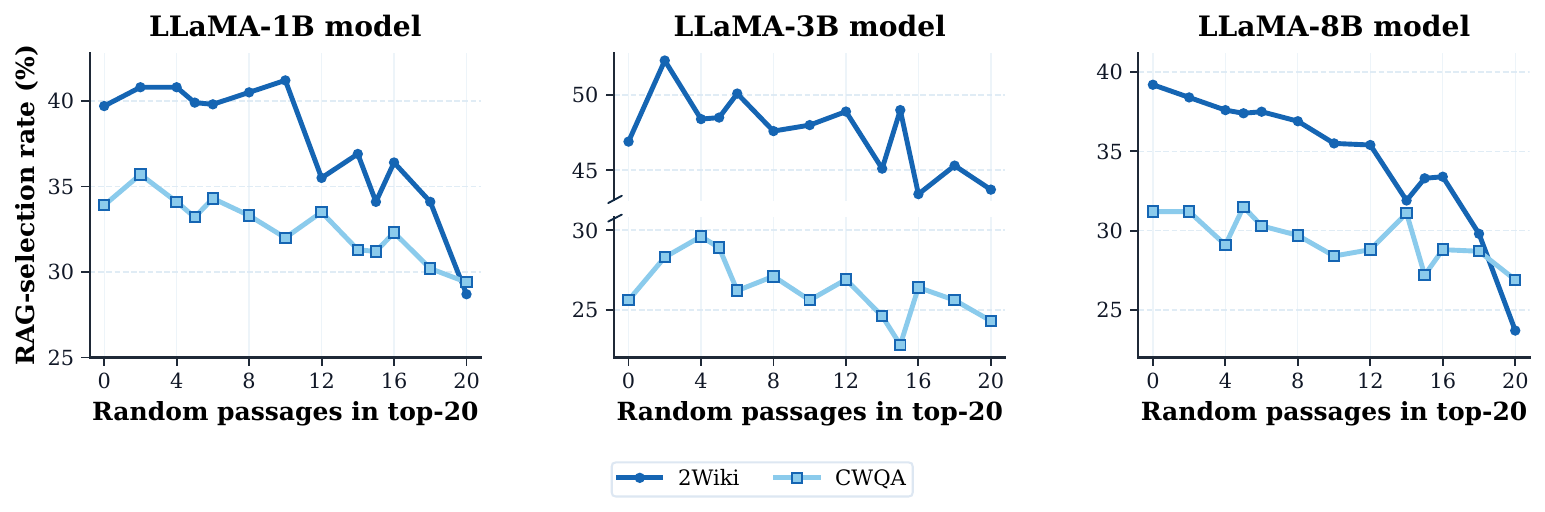}
\caption{
RAG-selection rate under retrieval corruption.
We replace different numbers of passages in the BM25 top-20 pool with random passages and measure how often \method{} selects the RAG answer.
Lower RAG selection under heavy corruption indicates that the evidence-binding margin helps \method{} back off from unreliable retrieval.
}
\label{fig:retrieval_noise_selection}
\end{figure}

\paragraph{Robustness to retrieval noise.}
Figure~\ref{fig:retrieval_noise_selection} corrupts the BM25 top-20 pool with random passages and tracks the RAG-selection rate.
As retrieval becomes less reliable, \method{} relies less on RAG.

The corruption experiment directly probes evidence binding.
\method{} observes no retrieval-quality labels and does not repair the retriever; it only detects when the RAG answer loses question-specific support under the scoring views.
The curves need not be monotonic, since small perturbations can remove distractors or leave the generated answer unchanged.
Under heavy corruption, \method{} backs off toward parametric memory.

\paragraph{Plug-and-play complementarity with Training-Free RAG pipelines.}
Table~\ref{tab:complementarity_delta_compact} tests \method{} on top of other training-free RAG pipelines when their RAG-side candidates remain distinct from Direct generation.
The gains are broadly positive, indicating that the rule is not tied to BM25-RAG alone.

\begin{table}[!t]
\centering
\scriptsize
\caption{
Complementarity with Training-Free RAG pipelines.
Each cell reports $\Delta$F1/$\Delta$EM after applying \method{} as an answer-level arbitration layer on top of the corresponding RAG pipeline.
The deltas are computed against the original pipeline without \method{} and averaged over \textsc{2WikiMQA} and \textsc{CWQA}.
Positive values indicate that source arbitration improves the pipeline.
}
\label{tab:complementarity_delta_compact}
\setlength{\tabcolsep}{3.5pt}
\renewcommand{\arraystretch}{1.05}
\begin{tabular}{@{}lccc@{}}
\toprule
RAG candidate
& LLaMA-1B
& LLaMA-3B
& LLaMA-8B \\
\midrule
IRCoT
& +2.23/+2.15
& +7.32/+6.45
& +6.08/+5.75 \\
FLARE
& +0.19/+0.35
& +0.34/+0.15
& $-$0.08/+0.00 \\
CLeHe-RAG
& +1.00/+1.05
& +4.68/+4.40
& +6.84/+4.80 \\
DTR-RAG
& +1.31/+1.30
& +2.96/+2.05
& +2.07/+1.15 \\
\bottomrule
\end{tabular}
\end{table}

The near-neutral FLARE result follows the same logic.
When the RAG-side answer is close to the model's native behavior, there is little candidate diversity for arbitration to exploit.
\method{} benefits from useful variation between Direct and RAG candidates; it does not create complementarity when the candidates collapse.
The division of labor is clear: RAG pipelines provide candidates, and \method{} decides when to trust them.

\paragraph{Takeaway.}
Direct and RAG are not globally ordered sources.
The better source changes with the question and retrieval quality.
\method{} makes this decision explicit: the prior margin checks memory, the binding margin checks evidence, and their sum defines a compact training-free boundary for deciding when retrieval should override memory.

\section{Conclusion}
\label{sec:conclusion}

We introduced \method{}, a training-free framework for answer-level source arbitration between closed-book and retrieval-augmented predictions.
Using only the frozen model's likelihoods, \method{} combines a parametric-prior margin with an evidence-binding margin to decide when retrieved evidence should override parametric memory.
Experiments on \textsc{2WikiMQA} and \textsc{CWQA} show consistent gains over Direct generation and BM25-RAG without fine-tuning, external judges, or additional generation.
Ablations and diagnostics show that the two margins capture complementary failure modes, that \method{} recovers part of the Direct/RAG oracle gap, and that the same rule remains useful for several training-free RAG pipelines when their candidates provide useful diversity.
The method is candidate-set bounded, but within that boundary it provides a lightweight and auditable reliability layer for RAG systems.

\section*{Impact Statement}

This work aims to improve the reliability of retrieval-augmented language-model systems through training-free answer-level source arbitration.
Because \method{} only selects between existing Direct and RAG candidates, its decisions can be audited through two likelihood margins rather than delegated to a new generator or judge.
The method may reduce cases where noisy retrieval overrides correct parametric memory.
However, it remains bounded by candidate quality and cannot ensure factual correctness when both candidates are wrong.
In high-stakes domains, \method{} should be used together with retrieval evaluation, answer verification, and domain-specific safety checks.

\FloatBarrier
\bibliography{references}
\bibliographystyle{icml2026}

\newpage
\suppressfloats[t]
\appendix

\section{Source-Selection Diagnostics}
\label{app:source_selection}

This appendix reports source-selection diagnostics in rate-only form.
The aligned candidate set used in the motivation analysis and main results is summarized by rates rather than row-level counts.

\begin{table}[H]
\centering
\scriptsize
\setlength{\tabcolsep}{4pt}
\renewcommand{\arraystretch}{1.05}
\caption{
Post-hoc source-selection rates in strict disagreement cases under the unified candidate-set definition.
$D{>}R{\to}D$ denotes the percentage of Direct-favored cases where \method{} selects Direct; $R{>}D{\to}R$ is defined analogously.
}
\label{tab:app_source_selection_rates}
\begin{tabular*}{\columnwidth}{@{\extracolsep{\fill}}lcc}
\toprule
Model & $D{>}R{\to}D$ & $R{>}D{\to}R$ \\
\midrule
LLaMA-1B & 60.59 & 66.94 \\
LLaMA-3B & 65.77 & 77.74 \\
LLaMA-8B & 60.00 & 76.39 \\
\bottomrule
\end{tabular*}
\end{table}

The strict disagreement rates show that \method{} is not simply increasing retrieval usage.
It selects Direct more often when Direct is better, and RAG more often when RAG is better.
The pattern is consistent with a source-reliability signal rather than a global preference for either source.

\FloatBarrier
\section{Plug-and-Play Complementarity with Training-Free RAG Pipelines}
\label{app:complementarity}

The main experiments use BM25-RAG as the retrieved-evidence candidate.
Here, we replace that candidate with the output of other training-free RAG pipelines and apply the same source decision rule.
To keep scoring comparable, \method{} evaluates each candidate against the same BM25 top-20 passage pool used in the main experiments.
The goal is to test whether the arbitration rule is specific to BM25-RAG or can serve as a post-hoc layer for other RAG candidates.

\begin{table}[H]
\centering
\scriptsize
\setlength{\tabcolsep}{2pt}
\renewcommand{\arraystretch}{1.03}
\caption{
Complementarity with Training-Free RAG pipelines on LLaMA-3.2-1B.
For each baseline, ``+ \method{}'' applies answer-level source arbitration between the Direct answer and the baseline's RAG-side answer.
Average is the arithmetic mean over \textsc{2WikiMQA} and \textsc{CWQA}; \textsc{TM} denotes \method{}.
}
\label{tab:app_complementarity_1b}
\begin{tabular*}{\columnwidth}{@{\extracolsep{\fill}}lcccccc}
\toprule
Method & 2W F1 & 2W EM & CW F1 & CW EM & Avg. F1 & Avg. EM \\
\midrule
IRCoT & 26.32 & 20.10 & 27.91 & 21.00 & 27.12 & 20.55 \\
IRCoT+\textsc{TM} & \textbf{27.45} & \textbf{21.30} & \textbf{31.25} & \textbf{24.10} & \textbf{29.35} & \textbf{22.70} \\
\midrule
FLARE & 22.77 & 16.60 & 28.74 & 21.50 & 25.75 & 19.05 \\
FLARE+\textsc{TM} & \textbf{23.19} & \textbf{17.10} & 28.68 & \textbf{21.70} & \textbf{25.94} & \textbf{19.40} \\
\midrule
CLeHe-RAG & 26.14 & 20.40 & 30.49 & 23.20 & 28.31 & 21.80 \\
CLeHe-RAG+\textsc{TM} & \textbf{26.44} & \textbf{20.90} & \textbf{32.18} & \textbf{24.80} & \textbf{29.31} & \textbf{22.85} \\
\midrule
DTR-RAG & 26.26 & 20.40 & 30.53 & 23.70 & 28.40 & 22.05 \\
DTR-RAG+\textsc{TM} & \textbf{26.66} & \textbf{21.00} & \textbf{32.76} & \textbf{25.70} & \textbf{29.71} & \textbf{23.35} \\
\bottomrule
\end{tabular*}
\end{table}

\begin{table}[!tb]
\centering
\scriptsize
\setlength{\tabcolsep}{2pt}
\renewcommand{\arraystretch}{1.03}
\caption{
Complementarity with Training-Free RAG pipelines on LLaMA-3.2-3B.
The setting is identical to Table~\ref{tab:app_complementarity_1b}.
}
\label{tab:app_complementarity_3b}
\begin{tabular*}{\columnwidth}{@{\extracolsep{\fill}}lcccccc}
\toprule
Method & 2W F1 & 2W EM & CW F1 & CW EM & Avg. F1 & Avg. EM \\
\midrule
IRCoT & 27.45 & 20.70 & 30.03 & 23.90 & 28.74 & 22.30 \\
IRCoT+\textsc{TM} & \textbf{31.35} & \textbf{24.50} & \textbf{40.77} & \textbf{33.00} & \textbf{36.06} & \textbf{28.75} \\
\midrule
FLARE & 24.04 & 17.90 & 38.27 & 30.20 & 31.16 & 24.05 \\
FLARE+\textsc{TM} & \textbf{24.29} & \textbf{18.00} & \textbf{38.72} & \textbf{30.40} & \textbf{31.50} & \textbf{24.20} \\
\midrule
CLeHe-RAG & 23.13 & 16.30 & 36.33 & 28.00 & 29.73 & 22.15 \\
CLeHe-RAG+\textsc{TM} & \textbf{27.36} & \textbf{20.60} & \textbf{41.47} & \textbf{32.50} & \textbf{34.41} & \textbf{26.55} \\
\midrule
DTR-RAG & 27.65 & 21.50 & 36.31 & 29.30 & 31.98 & 25.40 \\
DTR-RAG+\textsc{TM} & \textbf{30.06} & \textbf{23.30} & \textbf{39.82} & \textbf{31.60} & \textbf{34.94} & \textbf{27.45} \\
\bottomrule
\end{tabular*}
\end{table}

\begin{table}[!tb]
\centering
\scriptsize
\setlength{\tabcolsep}{2pt}
\renewcommand{\arraystretch}{1.03}
\caption{
Complementarity with Training-Free RAG pipelines on LLaMA-3.1-8B.
The setting is identical to Table~\ref{tab:app_complementarity_1b}.
}
\label{tab:app_complementarity_8b}
\begin{tabular*}{\columnwidth}{@{\extracolsep{\fill}}lcccccc}
\toprule
Method & 2W F1 & 2W EM & CW F1 & CW EM & Avg. F1 & Avg. EM \\
\midrule
IRCoT & 33.12 & 26.20 & 38.17 & 29.70 & 35.64 & 27.95 \\
IRCoT+\textsc{TM} & \textbf{38.23} & \textbf{31.70} & \textbf{45.21} & \textbf{35.70} & \textbf{41.72} & \textbf{33.70} \\
\midrule
FLARE & 31.61 & 24.90 & \textbf{43.74} & 34.10 & \textbf{37.67} & 29.50 \\
FLARE+\textsc{TM} & 31.50 & 24.90 & 43.68 & 34.10 & 37.59 & 29.50 \\
\midrule
CLeHe-RAG & 27.13 & 22.90 & 40.63 & 33.20 & 33.88 & 28.05 \\
CLeHe-RAG+\textsc{TM} & \textbf{34.94} & \textbf{28.80} & \textbf{46.49} & \textbf{36.90} & \textbf{40.72} & \textbf{32.85} \\
\midrule
DTR-RAG & 33.67 & 27.30 & 41.72 & 33.70 & 37.70 & 30.50 \\
DTR-RAG+\textsc{TM} & \textbf{35.18} & \textbf{28.20} & \textbf{44.36} & \textbf{35.10} & \textbf{39.77} & \textbf{31.65} \\
\bottomrule
\end{tabular*}
\end{table}

Across most baseline-model pairs, applying \method{} improves the final answer over the original RAG pipeline.
The gains depend on diversity between the Direct and RAG-side candidates.
The nearly neutral FLARE case illustrates the boundary: when the RAG-side answer is close to Direct behavior, there is little diversity for arbitration to exploit.

\section{Computation and Latency}
\label{app:time_efficiency}

\subsection{Computation Cost}
\label{app:time-computation-cost}

\method{} is training-free but not latency-free.
Its online cost includes Direct generation, RAG generation, and teacher-forced scoring for the two margins.
The numbers below are an implementation-level cost profile, not a theoretical lower bound.
The scoring passes are parallelizable, and Direct/RAG candidates or likelihood scores can be cached in offline evaluation and logging settings.

\subsection{Full Time Efficiency Results}
\label{app:time-efficiency-table}

Table~\ref{tab:time_efficiency_full} reports end-to-end seconds per query; lower is better.
For \method{}, latency includes Direct generation, BM25-RAG generation, and teacher-forced margin scoring.
The table should be read as a cost profile for an auditable arbitration layer, not as a claim of latency advantage.

\begin{table}[!tb]
\centering
\scriptsize
\setlength{\tabcolsep}{2.4pt}
\renewcommand{\arraystretch}{0.96}
\caption{
Compact time-efficiency results.
Latency is seconds per query; lower is better.
For \method{}, latency includes Direct generation, BM25-RAG generation, and teacher-forced margin scoring.
}
\label{tab:time_efficiency_full}
\begin{tabular*}{\columnwidth}{@{\extracolsep{\fill}}llccc}
\toprule
Model & Method & 2W & CW & Avg. \\
\midrule
\multirow{7}{*}{LLaMA-3.2-1B}
& Direct & \textbf{0.078} & \textbf{0.059} & \textbf{0.069} \\
& BM25-RAG & 0.287 & 0.279 & 0.283 \\
& IRCoT & 1.169 & 1.123 & 1.146 \\
& FLARE & \underline{0.176} & \underline{0.136} & \underline{0.156} \\
& CLeHe-RAG & 0.311 & 0.266 & 0.288 \\
& DTR-RAG & 0.443 & 0.420 & 0.431 \\
& \method{} & 1.436 & 1.408 & 1.422 \\
\midrule
\multirow{7}{*}{LLaMA-3.2-3B}
& Direct & \textbf{0.157} & \textbf{0.119} & \textbf{0.138} \\
& BM25-RAG & 0.535 & 0.511 & 0.523 \\
& IRCoT & 2.777 & 2.592 & 2.684 \\
& FLARE & \underline{0.282} & \underline{0.178} & \underline{0.230} \\
& CLeHe-RAG & 1.109 & 0.924 & 1.016 \\
& DTR-RAG & 0.710 & 0.501 & 0.605 \\
& \method{} & 2.581 & 2.518 & 2.550 \\
\midrule
\multirow{7}{*}{LLaMA-3.1-8B}
& Direct & \textbf{0.232} & \textbf{0.212} & \textbf{0.222} \\
& BM25-RAG & 1.012 & 0.990 & 1.001 \\
& IRCoT & 4.822 & 4.662 & 4.742 \\
& FLARE & \underline{0.356} & \underline{0.300} & \underline{0.328} \\
& CLeHe-RAG & 1.997 & 1.941 & 1.969 \\
& DTR-RAG & 0.828 & 0.741 & 0.784 \\
& \method{} & 4.651 & 4.611 & 4.631 \\
\bottomrule
\end{tabular*}
\end{table}

\method{} is not the fastest method because it uses both candidate generations and additional scoring.
Unlike iterative retrieval, however, its overhead is deterministic and can be amortized when candidates or likelihoods are cached.
It is therefore best viewed as a reliability layer for settings where both answers are already available or auditability matters more than minimal latency.

\section{Prompt Templates}
\label{app:prompt_templates}

This section describes the prompts used for answer generation and teacher-forced scoring.
The same short-answer style is used across Direct, RAG, and scoring prompts so that source decisions reflect evidence use rather than formatting differences.

\subsection{Context Construction}
\label{app:context_construction}

For each question, retrieved passages are concatenated in BM25 order.
Let $C(P)$ denote the resulting prompt context string constructed from the retrieved passage set $P$:
\begin{tcolorbox}[cosaPrompt,title=Concatenated Context]
\begin{Verbatim}[breaklines=true, fontsize=\footnotesize]
Passage 0: {passage_0}
Passage 1: {passage_1}
...
Passage 19: {passage_19}
\end{Verbatim}
\end{tcolorbox}

The same context string is used for RAG answer generation, question-context likelihood scoring, and context-only likelihood scoring.

\subsection{Question-Only Prompt}
\label{app:question_only_prompt}

The question-only prompt is used for Direct generation and for computing the prior margin $\mprior$.

\begin{tcolorbox}[cosaPrompt,title=Question-only Prompt]
\begin{Verbatim}[breaklines=true, fontsize=\footnotesize]
You are a helpful AI assistant.
Answer the user's question concisely with a short phrase or a single word.

Question:
{question}

Answer:
\end{Verbatim}
\end{tcolorbox}

\subsection{Question-Context Prompt}
\label{app:question_context_prompt}

The question-context prompt is used for BM25-RAG generation and for the question-context term in the binding score.

\begin{tcolorbox}[cosaPrompt,title=Question-Context Prompt]
\begin{Verbatim}[breaklines=true, fontsize=\footnotesize]
####CONTEXT begin####
{context}
####CONTEXT end####

You are a helpful AI assistant.
Answer the user's question according to the CONTEXT.
Your answer should be concise: a short phrase or a single word.

Question:
{question}

Answer:
\end{Verbatim}
\end{tcolorbox}

\subsection{Context-Only Scoring Prompt}
\label{app:context_only_prompt}

The context-only prompt removes the question while preserving the retrieved passages.
It estimates passage-only salience and is used in the context-only term of the evidence-binding margin.

\begin{tcolorbox}[cosaPrompt,title=Context-only Scoring Prompt]
\begin{Verbatim}[breaklines=true, fontsize=\footnotesize]
####CONTEXT begin####
{context}
####CONTEXT end####

You are a helpful AI assistant.
Produce the concise answer phrase that is most supported by the CONTEXT.

Answer:
\end{Verbatim}
\end{tcolorbox}

\subsection{Teacher-Forced Candidate Scoring}
\label{app:teacher_forced_scoring}

For likelihood scoring, the prompt prefix is fixed and the candidate answer is teacher-forced.
All likelihoods are normalized by answer token length.

\begin{tcolorbox}[cosaPrompt,title=Teacher-forced Scoring]
\begin{Verbatim}[breaklines=true, fontsize=\footnotesize]
Prompt prefix:
{one of the prompts above}

Candidate answer:
{candidate_answer}

Score:
length-normalized log likelihood of {candidate_answer}
\end{Verbatim}
\end{tcolorbox}

For each example, the placeholders are instantiated as follows:
\begin{itemize}[leftmargin=*]
\item Insert the user question into \verb|{question}|.
\item Insert the concatenated top-20 BM25 passages into \verb|{context}|.
\item Insert either $y_D$ or $y_R$ into \verb|{candidate_answer}| during teacher-forced scoring.
\end{itemize}

\end{document}